\documentclass{article}
\usepackage{ijcai19}

\usepackage{times}
\usepackage{soul}
\usepackage{url}
\usepackage[hidelinks]{hyperref}
\usepackage[utf8]{inputenc}
\usepackage[small]{caption}
\usepackage{graphicx}
\usepackage{amsmath}
\usepackage{booktabs}
\usepackage{algorithm}
\usepackage{algorithmic}
\usepackage{color,xcolor}
\usepackage{subfigure}
\usepackage{amsfonts}
\usepackage{multirow}

\title{From Words to Sentences: A Progressive Learning Approach for \\Zero-resource Machine Translation with Visual Pivots}

\author{
Shizhe Chen$^1$\footnote{This work was performed when Shizhe Chen was visiting Microsoft Research Asia as a research intern.}
\and
Qin Jin$^1$\thanks{Qin Jin is the corresponding author.} 
\And
Jianlong Fu$^{2}$
\affiliations
$^1$Renmin University of China, Beijing, P.R. China\\
$^2$Microsoft Research Asia, Beijing, P.R. China
\emails
\{cszhe1, qjin\}@ruc.edu.cn,
jianf@microsoft.com
}

\begin{document}

\maketitle

\begin{abstract}
	The neural machine translation model has suffered from the lack of large-scale parallel corpora.
	In contrast, we humans can learn multi-lingual translations even without parallel texts by referring our languages to the external world.
	To mimic such human learning behavior, we employ images as pivots to enable zero-resource translation learning.
	However, a picture tells a thousand words, which makes multi-lingual sentences pivoted by the same image noisy as mutual translations and thus hinders the translation model learning.
	In this work, we propose a progressive learning approach for image-pivoted zero-resource machine translation.
	Since words are less diverse when grounded in the image, we first learn word-level translation with image pivots, and then progress to learn the sentence-level translation by utilizing the learned word translation to suppress noises in image-pivoted multi-lingual sentences.
	Experimental results on two widely used image-pivot translation datasets, IAPR-TC12 and Multi30k, show that the proposed approach significantly outperforms other state-of-the-art methods.
\end{abstract}

\section{Introduction}

The recent success of neural machine translation (NMT) \cite{bahdanau2014neural} has greatly benefited from large-scale high-quality parallel corpora.
However, such NMT models are data-hungry and perform poorly without sufficient parallel data \cite{zoph2016transfer}. 
Due to the high expense of collecting parallel texts, more and more researchers are paying attention to develop NMT models under the zero-resource condition where no parallel source-target texts are available.


Inspired by how we humans learn a novel language when no parallel texts are available, for example connecting sentences in two languages that describe the same image, researchers have proposed to employ images as pivots for zero-resource machine translation, which can benefit from abundant images with mono-lingual descriptions on the Internet \cite{sharma2018conceptual}.
The shared principle in previous image-pivoted works \cite{nakayama2017zero,lee2017emergent,chen2018zero} assumes that the source sentence and the target sentence are semantically equivalent as they are describing the same image.

\begin{figure} \centering
	\includegraphics[width=1 \linewidth]{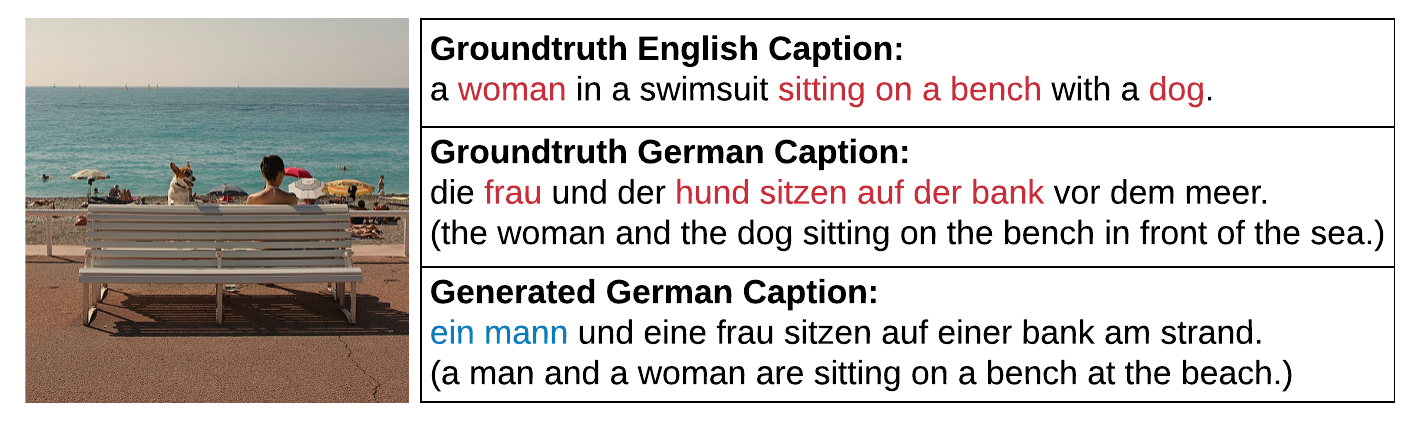}
    \caption{A picture tells a thousand words - illustration of the challenge in image-pivoted zero-resource machine translation. Only few words in red are correct translations even in groundtruth captions of the same image. Captioning mistakes in blue can further make such multi-lingual captions noisy as mutual translation. We translate German captions in brackets for non-German readers.}
	\label{fig:intro_challenge}
\end{figure}

However, due to the one-to-many relationship between images and captions (as known as \textit{``a picture tells a thousand words"}), multi-lingual sentences describing the same image are not necessarily good mutual translations.
As shown in Figure~\ref{fig:intro_challenge}, although the captions in English and German both accurately describe the image, they capture different aspects in the image and only few words in the captions are correct mutual translations.
Moreover, since images with multi-lingual captions are hard to obtain in realistic settings, an image caption model is usually adopted to generate multi-lingual caption sentences for the image.
However, due to the imperfection of caption models, the semantic discrepancy of generated multi-lingual captions could be even larger.

In this work, we propose a progressive learning approach to overcome above challenges in the image-pivoted zero-resource machine translation.
We propose to learn the translation in an easy-to-advanced progressive way, by firstly grasping the word-level translation with the help of image pivots and then progressing to learn more challenging sentence-level translation with the assistance of word translation and image pivots. 
To be specific, since words grounded in certain regions of the image are less diverse than sentences, word-level translation can be more effectively learned based on image pivots.
The multi-lingual word representations learned from the word translation and the image pivots altogether are used to enhance the sentence translation from two aspects:
i) suppressing noises in image-pivoted multi-lingual sentences via re-weighting sentences at fine-grained token-level;
ii) supporting the learning of language-agnostic sentence representation for cross language decoding via auto-encoding.
The two aspects are complementary to train the NMT model.
We carry out extensive experiments on two benchmark image-pivot machine translation datasets: IAPR-TC12 and Multi30k.
Our proposed approach significantly outperforms other state-of-the-art image-pivot methods.


\section{Related Work}

The encoder-decoder based neural machine translation (NMT) model \cite{cho2014properties,bahdanau2014neural} has achieved great success in recent years.
However, it requires large-scale parallel texts for training, which performs poorly without sufficient data \cite{zoph2016transfer,castilho2017neural}.
There are mainly three types of methods to avoid the reliance on source-target parallel data, namely third-language pivot, mono-lingual based and visual pivot methods.

The third-language pivot methods \cite{johnson2017google,chen2017teacher} demand the source-to-pivot and pivot-to-target parallel corpus to enable zero-resource translation from source to target.
\cite{johnson2017google} trains a universal encoder-decoder with multiple language pairs, which can perform well in novel language combinations.
However, it is not trivial to obtain the pivot language parallel data.

The recent mono-lingual based methods \cite{artetxe2017unsupervised,lample2017unsupervised} only utilize large-scale mono-lingual corpora for translation.
\cite{lample2018phrase} summarizes three key elements for mono-lingual based methods: careful initialization, strong language model, and back-translation, which achieved promising results for both phrase-based and NMT models.
The reason we use image-pivot for translation is that images can help reduce ambiguity of texts especially for visual-related sentences such as commercial product descriptions \cite{zhou2018visual,calixto2017using}.

The image-pivot approaches leverage images to connect unpaired source and target languages.
\cite{kiela2015visual,chen2019unsupervised} have shown the effectiveness of image pivots for bilingual lexicon induction.
\cite{su2018unsupervised} follow mono-lingual based methods but utilize images to enhance decoding performance.
The 3-way model \cite{nakayama2017zero} maps source sentences and images into common space and employs an image-to-target caption model for translation.
However, it cannot embrace attention mechanism and results in noisy translations since information in images and sentences is not equal. 
\cite{chen2018zero,lee2017emergent} propose a multi-agent communication game with a captioner and translator. The captioner generates source sentence to describe an image, and the translator is trained to maximize rewards from relevancy between image and translated target sentence.
\cite{chen2018zero} utilizes log-likelihood as the reward while \cite{lee2017emergent} utilizes image retrieval performance.
However, since captions related to images are not necessarily good mutual translations, such learning approaches also suffer from noisy rewards.
In this work, we propose to suppress such noises and progressively learn the translation in an easy-to-advanced way.
\section{The Proposed Approach}
The goal of zero-resource machine translation is to learn source to target translation without any source-target sentence pairs.  We propose to utilize images as pivots to enable zero-resource machine translation. 
Assume we have two mono-lingual image caption datasets: $D_x=\{(I_{x}^{(i)}, X^{(i)})\}_{i=1}^{N_x}$ in the source language and $D_y=\{(I_{y}^{(i)}, Y^{(i)})\}_{i=1}^{N_y}$ in the target language, 
where $I$ denotes the image, and captions $X$ and $Y$ consist of word sequences $\{x_{1}, \cdots, x_{T_x} \}$ and $\{y_{1}, \cdots, y_{T_y} \}$ respectively.
We omit the superscript $i$ for simplicity hereinafter.
The image sets $I_x$ and $I_y$ do not overlap, which means that an image has only one caption, either in source language or target language.
The images are used as pivots during the training stage, but are not involved during the test stage.

\begin{figure}
    \centering
	\includegraphics[width=1 \linewidth]{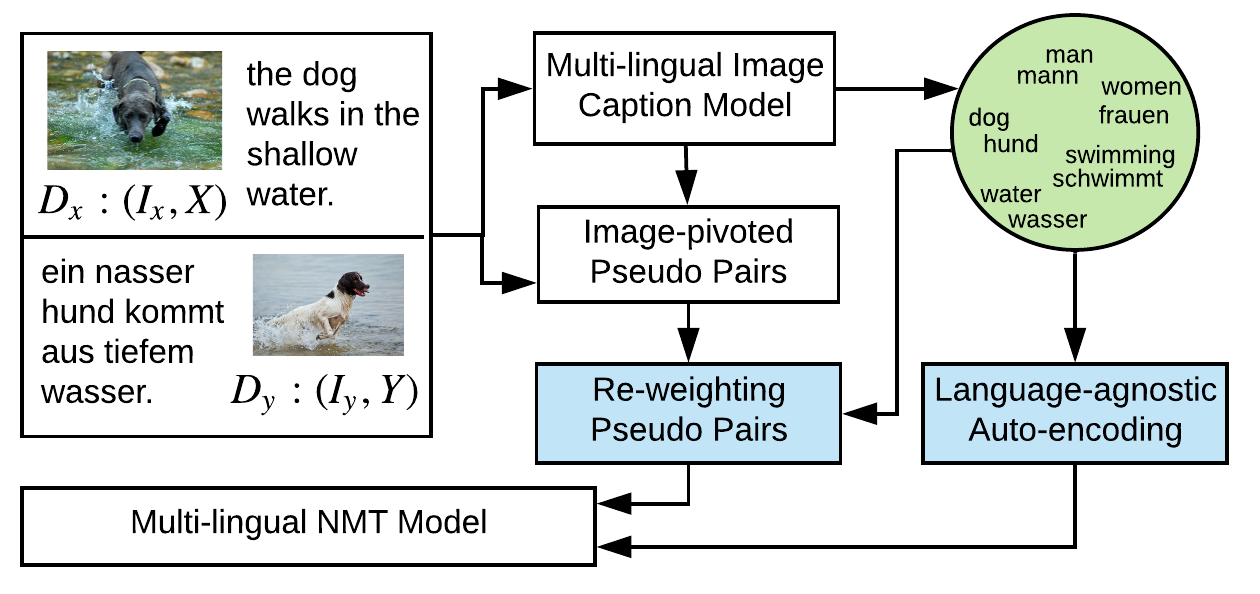}
	\caption{The overall progressive learning framework for image-pivoted zero-resource machine translation. 
	We firstly learn word translations from image pivot in the green module and then we advance to more challenging sentence translation in blue modules.
	}
	\label{fig:system_framework}
\end{figure}

Figure~\ref{fig:system_framework} illustrates the proposed progressive learning approach for image-pivoted zero-resource machine translation.
Firstly, from the mono-lingual image caption datasets $D_x$ and $D_y$, we can build image caption models $f_{i \rightarrow x}$ and $f_{i \rightarrow y}$ to translate an image into sentence descriptions in source and target languages respectively. 
Therefore, for each image $I_x \in D_x$, we can obtain triplet captions $(X, \tilde{X}, \tilde{Y})$ where $X$ refers to the groundtruth caption, $\tilde{X}$ and $\tilde{Y}$ refer to the generated captions from the image caption models $f_{i \rightarrow x}$ and $f_{i \rightarrow y}$ in source and target languages respectively.
Similarly, for image $I_y \in D_y$, we can obtain triplet captions $(Y, \tilde{X}, \tilde{Y})$.
We then induce $(X, \tilde{Y})$, $(\tilde{X}, Y)$ and $(\tilde{X}, \tilde{Y})$ from above triplets as image-pivoted source-target pseudo sentence pairs, which can be utilized to train NMT models.

However, since the diversity of descriptive sentences is quite large, such pseudo pair may not be precisely semantic matched, which would greatly hinder the translation performance.
In order to suppress noises in pseudo pairs, we propose a progressive learning approach which firstly learns word translation from image pivots.
Since the diversity of words is better constrained when the word is grounded in image regions, word translation can be more effectively learned with the image-pivot approach. 
The word translation encodes multi-lingual words into a common semantic space,
which is then utilized to assist image-pivoted sentence translation from two aspects: 1) to re-weight image-pivoted pseudo pairs and 2) to learn language-agnostic sentence representations.
In such progressive manner, the NMT model is able to alleviate noises from image-pivot learning.

In the following, we will firstly describe the structure of NMT model in Section~\ref{sec:nmt_model}, and then introduce the progressive learning strategy to train the NMT model with image pivots in details in Section~\ref{sec:progressive_train}.

\begin{figure} \centering
	\includegraphics[width=1 \linewidth]{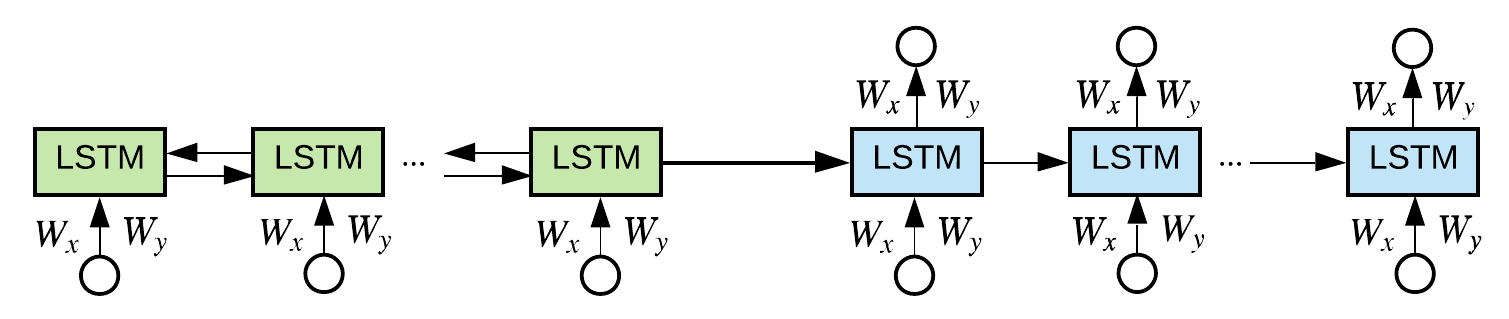}
	\caption{The structure of the NMT model. The encoder and decoder both contain source and target embedding matrices $W_x$ and $W_y$, so we can encode and decode source and target sentences in one model.}
	\label{fig:nmt_model}
\end{figure}

\subsection{Neural Machine Translation Model}
\label{sec:nmt_model}
We utilize a shared multi-lingual encoder-decoder architecture as our NMT model, which can perform source-to-target and target-to-source translation in one model.
Given the source sentence $X=\{x_1,...,x_{T_x}\}$ and the target sentence $Y=\{y_1,...,y_{T_y}\}$,
the encoder converts the source sentence into a sequence of vectors, and then the decoder sequentially predicts target word $y_t$ conditioning on the encoded vectors and previously generated words $y_{<t}$.

Specifically, our encoder is a bi-directional LSTM \cite{hochreiter1997long} with a source word embedding matrix $W_x$ and a target word embedding matrix $W_y$.
The source and target sentences share the same encoder parameters except the word embedding matrix as follows:
\begin{eqnarray}
    z^x_t = \mathrm{biLSTM}(W_x x_t, z^x_{t-1}, z^x_{t+1}; \Theta_e)\\
    z^y_t = \mathrm{biLSTM}(W_y y_t, z^y_{t-1}, z^y_{t+1}; \Theta_e)
\end{eqnarray}
where $\Theta_e$ are parameters of the bi-directional LSTM.

The decoder is a LSTM with both source word embedding $W_x$ and target word embedding $W_y$.
Suppose $z=\{z_1, \cdots, z_T\}$ is the encoded input sentence, 
the decoder can translate $z$ into the source sentence $X$ by:
\begin{eqnarray}
\label{eqn:decoder}
   p(x_t|x_{<t}, z) = \mathrm{softmax}( W_{x} h_{t})\\
   h_{t} = \mathrm{LSTM}([W_x x_{t-1}, c_{t}], h_{t-1}; \Theta_d)
\end{eqnarray}
where $\Theta_d$ are parameters in the decoder LSTM, $h_0$ is initialized as $z_{T}$, $[\cdot]$ is the vector concatenation operation and $c_t$ is a context vector to employ relevant input vector $z_i$ to predict the target word $x_t$ via attention mechanism \cite{bahdanau2014neural}.
The computation of $c_t$ is formulated as:
\begin{eqnarray}
c_t &=& \sum_{i=1}^{T} a_{i, t} z_i\\
a_{i, t} &=& \frac{\mathrm{exp}(f_a ([h_{t-1}, z_i]))}{\sum_j \mathrm{exp}(f_a ([h_{t-1}, z_j]))} ,
\end{eqnarray}
where $f_a$ is a feed forward neural network to compute the attention weight $a_i$ for each $h_i$.
Therefore, the probability of generating $X$ conditioning on $z$ is:
\begin{equation}
p(X|z) = \prod_{t=1}^{T_x} p(x_t|x_{<t}, z)
\end{equation}
Similarly, the decoder can also translate the input $z$ into the target sentence $Y$ by replacing $W_x$ in Eq (3) and Eq (4) with $W_y$.
Figure~\ref{fig:nmt_model} presents the structure of the NMT model.


\subsection{Progressive Learning}
\label{sec:progressive_train}
In this section, we describe in details the progressive learning procedure, from learning the word-level translation to more challenging sentence-level translation. 

\subsubsection{Learning Word-level Translation}
\label{sec:word_translation}
In order to translate multi-lingual words, similar to work in \cite{chen2019unsupervised}, we build a multi-lingual image caption model to encode the source and target words into a joint semantic space.
The multi-lingual image caption model is based on the encoder-decoder framework, where the encoder converts an image into a set of visual features, and the decoder generates sentences conditioning on the visual features with attention mechanism. 
For image captioning in the source and target languages, the encoder and decoder are shared except the word embedding matrices.
Therefore, the word embedding matrices of the source and target languages are enforced to be in a common space constrained by the image pivots.
We denote the learned word embedding matrices for the source and target words as $W_x$ and $W_y$, which are employed in our NMT model and remain fixed.

\subsubsection{Re-weighting Image-pivoted Pseudo Sentence Pairs}
\label{sec:token_reweight}

\begin{figure} \centering
	\subfigure[Re-weighting image-pivoted pseudo sentence pairs. 
	] { \centering
		\label{fig:token_reweight}
		\includegraphics[width=0.9\linewidth]{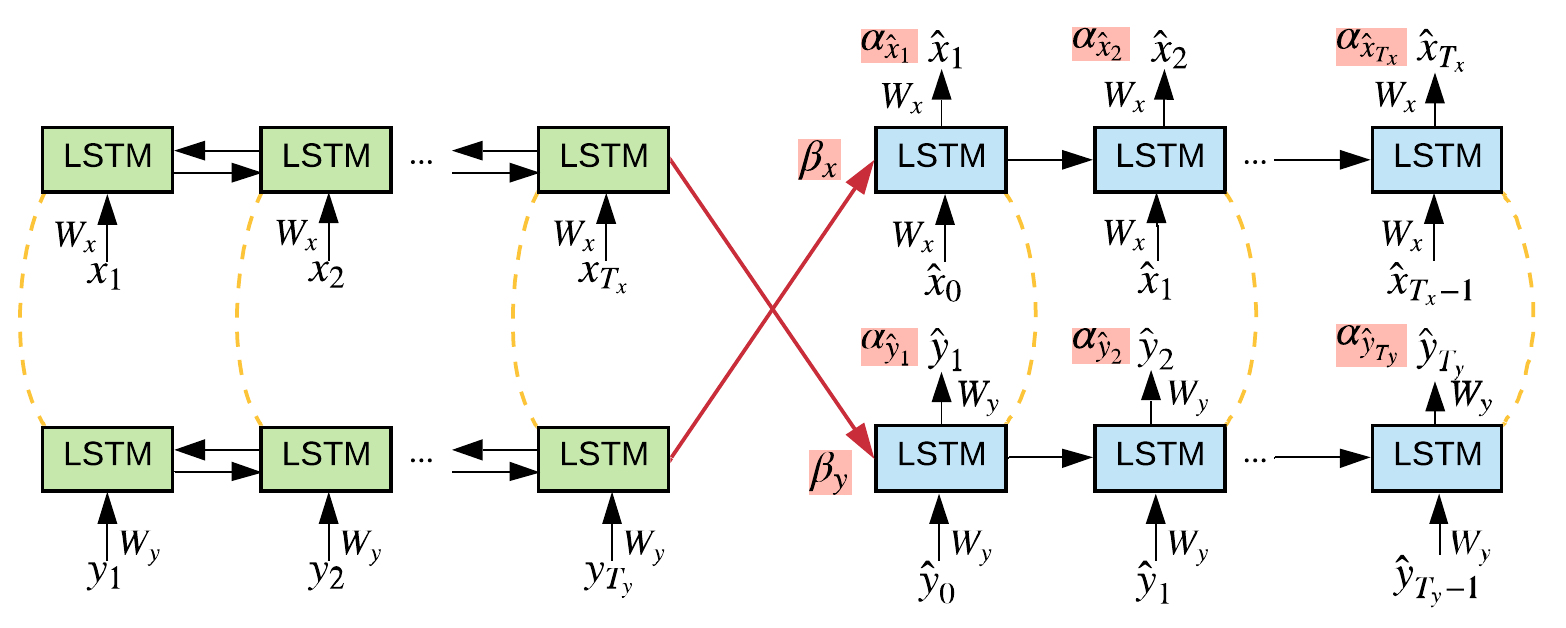}
	} 
	\subfigure[Language-agnostic auto-encoding. ] { \centering
		\label{fig:auto_encoding}
		\includegraphics[width=0.9\linewidth]{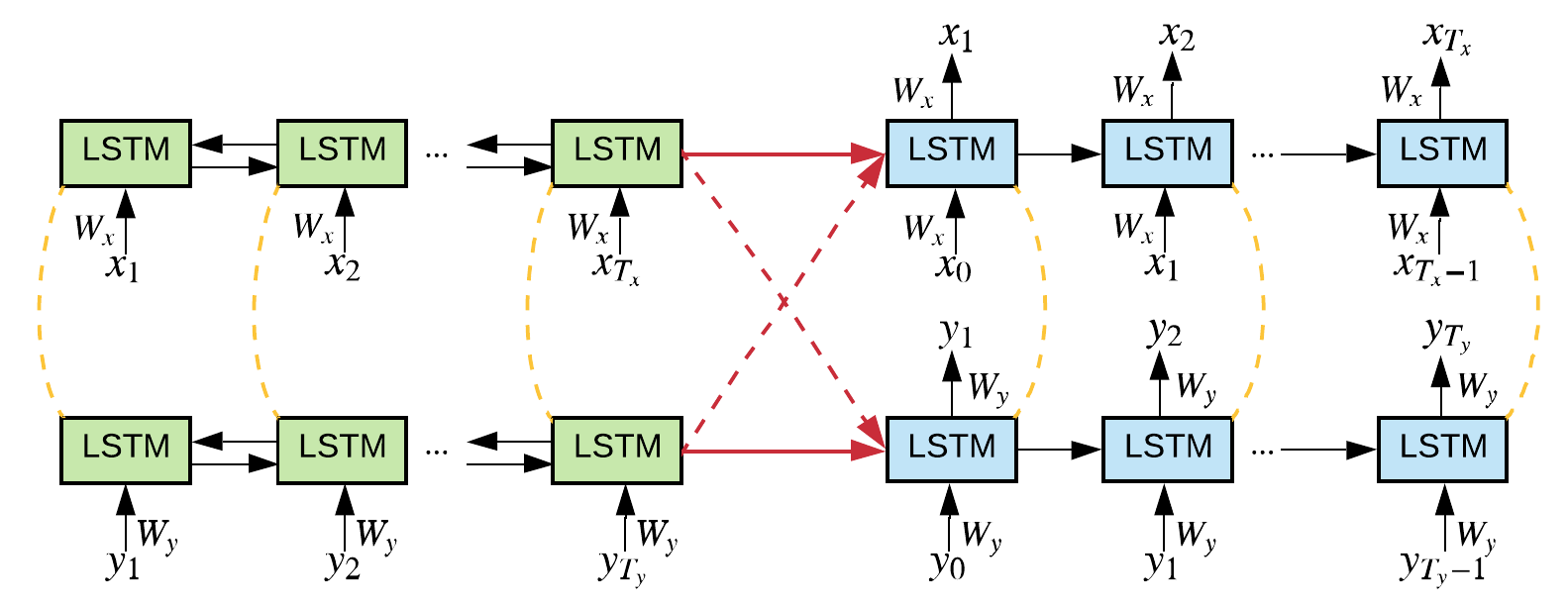}
	}
	\caption{Training of the NMT model. The red solid lines denote translation in training; red dashed lines denote implicitly learned cross-language translation; yellow dashes lines denote tied weights.}
	\label{fig:nmt_train}
\end{figure}

We employ above learned $W_x, W_y$ to assess the qualities of image-pivoted pseudo sentence pairs.
We formulate the semantic distance of the sentence pair as a special case of Earth Mover's Distance (EMD) \cite{rubner2000earth,kusner2015word}, which is to find the minimal transportation solution from one sentence to another.
For source sentence $X=\{x_1, \cdots, x_{T_x}\}$ and target sentence $Y=\{y_1, \cdots, y_{T_y}\}$, their EMD distance $d$ is:
\begin{equation}
\label{eqn:emd_optimization}
\begin{aligned}
& d = \underset{\mathrm{A} \ge 0 }{\text{min}}
& & \sum_{i, j} \mathrm{A}_{i, j} \ D_{i, j} \\
& \text{subject to:} 
& & \sum_j \mathrm{A}_{i, j} = 1 / {T_x}, \; i = 1, \ldots, T_x \\
& & & \sum_i \mathrm{A}_{i, j} = 1 / {T_y}, \; j = 1, \ldots, T_y \\
\end{aligned}
\end{equation}
where $D_{i, j}$ is the cosine distance of $x_i$ and $y_j$ based on $W_x$ and $W_y$.
The optimal matrix $\mathrm{A}$ can be solved efficiently via an off-the-shelf toolkit proposed by \cite{pele2009}, which can be viewed as words alignment in the pseudo pair. 
Besides the sentence distance $d$, we could also derive token-level distances $d_{x_i}$ and $d_{y_j}$ for $x_i$ and $y_j$ respectively based on the optimal $\mathrm{A}$, where $d_{x_i}=\sum_{j} \mathrm{A}_{i, j} \ D_{i, j}$ and $d_{y_j}=\sum_{i} \mathrm{A}_{i, j} \ D_{i, j}$.
Based on the sentence-level and token-level distance, we propose to re-weight the $i$-th pseudo pair at both sentence-level and fine-grained token-level via the inverse distance weight as follows:
\begin{eqnarray}
	\alpha_{x_t}^{(i)} &=& \frac{1}{(1 + d_{x_t}^{(i)} - \mathrm{min}_{k} d_{x_k}^{(i)})^{\lambda_{token}}} \\
	\alpha_{y_t}^{(i)} &=& \frac{1}{(1 + d_{y_t}^{(i)} - \mathrm{min}_{k} d_{y_k}^{(i)})^{\lambda_{token}} } \\
	\beta^{(i)} &=& \frac{1}{(1 + d^{(i)} - \mathrm{min}_{k} d^{(k)})^{\lambda_{sent}} }
\end{eqnarray}
where $\lambda_{token}, \lambda_{sent}$ are hyper-parameters to punish noisy tokens and sentences.
So $\alpha_{x_t}^{(i)}$ and $\alpha_{y_t}^{(i)}$ represent the relative importance of token $x_t$ and $y_t$ in the source and target sentence to form a good mutual translation pair, and $\beta^{(i)}$ is the relative quality of pair $i$ among all pseudo pairs.
Figure~\ref{fig:token_reweight} illustrates the re-weighting training process.
Therefore, the objective function for image-pivoted sentence pairs is:
\begin{equation}
\begin{aligned}
\label{eqn:sentence_loss}
L_{pivot} = - \sum_{i=1}^{N} \beta^{(i)} \sum_{t=1}^{T_y} \alpha_{y_t}^{(i)} \mathrm{log}\ p(y^{(i)}_t|y^{(i)}_{<t}, z^{x^{(i)}}) \\ - \sum_{i=1}^{N} \beta^{(i)} \sum_{t=1}^{T_x} \alpha_{x_t}^{(i)} \mathrm{log}\ p(x^{(i)}_t|x^{(i)}_{<t}, z^{y^{(i)}})
\end{aligned}
\end{equation}

\subsubsection{Language-agnostic Auto-encoding}
\label{sec:auto_encoding}

We propose to employ auto-encoding to learn a language-agnostic sentence representation for cross language translation as illustrated in Figure~\ref{fig:auto_encoding}.
We fix source and target word embedding matrices in our NMT model with $W_x$ and $W_y$ which are in the common space, so that the sentence representation encoded by the shared encoder are constrained in a common latent space.
Since it is trivial to auto-encode a sentence, we apply the corruption operation $C(\cdot)$ on the original sentence as \cite{lample2017unsupervised}, which includes word order jitter, word insertion and deletion.
The corrupted sentence is used as the input and the NMT model is trained to reconstruct the original sentence from the corrupted one.
We apply the denoising auto-encoding for both source and target languages, so the objective is to minimize:
\begin{equation}
\begin{aligned}
L_{ae} = - \sum_{i=1}^{N_x} \mathrm{log} \  p(X^{(i)}|C(X^{(i)})) \\
			 -  \sum_{i=1}^{N_y} \mathrm{log} \ p(Y^{(i)}|C(Y^{(i)}))
\end{aligned}
\end{equation}

The full objective function to train the NMT model is:
\begin{equation}
\label{eqn:loss}
L = L_{pivot} + \lambda L_{ae}
\end{equation}
where $\lambda$ is the weight to balance the two losses. 


 \begin{table}
 	\begin{tabular}{c|cc|cc|c} \toprule
 		& \multicolumn{2}{|c|}{Train} & \multicolumn{2}{c|}{Val} & Test \\ 
 		Pairs & I-En & I-De & I-En & I-De & En-De \\ \midrule
 		\small IAPR-TC12 & 9k & 9k & 500 & 500 & 1k \\
 		\small Multi30k & 14.5k & 14.5k & 507 & 507 & 1k \\ \bottomrule
 	\end{tabular}
 	\caption{Data splits for the IAPR-TC12 and Multi30k datasets.}
 	\label{tab:dataset}
 \end{table}
 
\section{Experiments}
\subsection{Experimental Setup}
We utilize two benchmark image-pivot translation datasets IAPR-TC12 and Multi30k.
The IAPR-TC12 dataset \cite{grubinger2006iapr} contains 20K images and each image is annotated with multi-sentences in English and its translation in German.
We follow \cite{chen2018zero} to use the first sentence since it describes the most salient image content.
We randomly select 18K images for training, 1K for validation and 1K for testing.
Since in the realistic image-pivot setting, images in different languages are mostly non-overlapped, we randomly split the training and validation set into two parts of equal size.
One part is constructed only with image-English pairs and the other only with image-German pairs.
The Multi30k dataset \cite{elliott2016multi30k} contains 30K images with two task annotations, one for machine translation and the other for multi-lingual captioning.
We utilize the former task annotation, where each image is annotated with one English description and its German translation.
We follow the standard split in \cite{nakayama2017zero} with 29K, 1,014 and 1K in the training, validation and test sets respectively.
We also apply the similar non-overlapping split operation as in IAPR-TC12 to simulate the non-overlap setting.
Table~\ref{tab:dataset} presents the data split in our experiments.

We use Moses SMT Toolkit \cite{koehn2007moses} to normalize and tokenize descriptions.
For IAPR-TC12 dataset, we keep words appeared more than 3 times, which results in 1,621 words for English and 2,102 words for German.
For Multi30K dataset, we employ a joint byte pair (BPE) \cite{sennrich2016neural} with 10k merge operations, which results in 5,202 tokens for English and 7,065 tokens for German.
 
For the multi-lingual image caption model, we leverage the Resnet152 pretrained on the ImageNet \cite{he2016deep} as the encoder and a single-layer LSTM with 512 hidden units as the decoder.
We utilize beam search with beam width of 5 to generate one description for each image.
For the NMT model, the encoder is a one-layer bidirectional LSTM with 256 hidden units and the decoder is a one-layer LSTM with 512 hidden units.
We set hyper-parameters $\lambda_{token}=10, \lambda_{sent}=5$ and $\lambda=1$ based on validation performance.
We utilize the Adam algorithm to train models with learning rate of 0.0005 and batch size of 64.
The best model is selected by the loss on validation set.
We evaluate the machine translation performance with BLEU4 metric \cite{papineni2002bleu}.

\begin{table}
\centering
\begin{tabular}{lc|cc|cc} \toprule
\multicolumn{2}{l|}{\multirow{2}{*}{}} & \multicolumn{2}{c}{IAPR-TC12} & \multicolumn{2}{c}{Multi30k} \\
\multicolumn{2}{l|}{} & De-En & En-De & De-En & En-De \\ \midrule
\multicolumn{2}{l|}{3-way model} & 13.9 & 8.6 & 8.4 & 8.0 \\
\multicolumn{2}{l|}{Multi-agents\footnotemark[1]} & 18.6 & 14.2 & - & - \\
\multicolumn{2}{l|}{Emergent model\footnotemark[2]} & - & - & 6.5 & 7.4 \\ 
\multicolumn{2}{l|}{S-txt-img\footnotemark[2]} & - & - & 7.5 & 7.7 \\ \midrule
\multicolumn{1}{l|}{\multirow{3}{*}{ours}} & $L_{pivot}$ & 38.0 & 30.3 & 9.9 & 8.4  \\
\multicolumn{1}{l|}{} & $L_{ae}$ & 58.9 & 39.8 & 21.9 & 17.6 \\
\multicolumn{1}{l|}{} & $L_{pivot}+L_{ae}$ &\textbf{61.3} & \textbf{47.1} & \textbf{23.0} & \textbf{18.3} \\ \bottomrule
\end{tabular}
\caption{The BLEU4 performance of different methods for image-pivoted zero-resource machine translation.}
\label{tab:sota_comparison}
\end{table}

\footnotetext[1]{Multi-agents method \cite{chen2018zero} utilizes different training and evaluation setup on Multi30k dataset compared with others.}
\footnotetext[2]{The two works did not report results on the IAPR-TC12 dataset.}

\begin{table}
\centering
\begin{tabular}{cc|cc|cc} \toprule
\multicolumn{2}{c|}{\multirow{2}{*}{}} & \multicolumn{2}{c}{IAPR-TC12} & \multicolumn{2}{c}{Multi30k} \\ 
\multicolumn{2}{c|}{} & De-En & En-De & De-En & En-De \\ \midrule
\multicolumn{1}{c|}{\multirow{2}{*}{$L_{pivot}$}} & w/o & 31.0 & 25.5 & 8.5 & 7.7 \\ 
\multicolumn{1}{c|}{} & with & \textbf{38.0} & \textbf{30.3} & \textbf{9.9} & \textbf{8.4} \\  \midrule
\multicolumn{1}{c|}{\multirow{2}{*}{$L_{ae}$}} & w/o & 39.6 & 26.8 & 7.9 & 6.8 \\
\multicolumn{1}{c|}{} & with & \textbf{58.9} & \textbf{39.8} & \textbf{21.9} & \textbf{17.6} \\ \bottomrule
\end{tabular}
\caption{Translation performance comparison of without and with the progressive learning approach for different sentence losses.}
\label{tab:ablation_performance}
\end{table}

\subsection{Comparison with State-of-the-Art Methods}
We compare the proposed progressively learned NMT model with state-of-the-art image-pivoted models as follows:
\begin{enumerate}
\item 3-way model \cite{nakayama2017zero}: It utilizes an target image caption model to translate a learned modality-agnostic feature of the source sentence.
\item Multi-agents \cite{chen2018zero}: The NMT model is trained to translate a generated source caption of $I_y$ with image-relevance rewards from the log probability of predicting groundtruth $Y$.
\item Emergent model \cite{lee2017emergent}: Similar to \cite{chen2018zero}, but the model utilizes an image retrieval task to calculate the image-caption relevance as rewards.
\item S-txt-img \cite{su2018unsupervised}: The model is similar to mono-lingual based methods with auto-encoding and cycle-consistency loss, but employs an additional image encoder in training which can be absent in testing.
\end{enumerate}

\begin{table*}[htb] \centering
	\begin{tabular}{cll} \toprule
		Image & Source and groundtruth target sentence & Translated sentences \\ \midrule
		
		\begin{minipage}{.12\textwidth}
			\includegraphics[width=\linewidth]{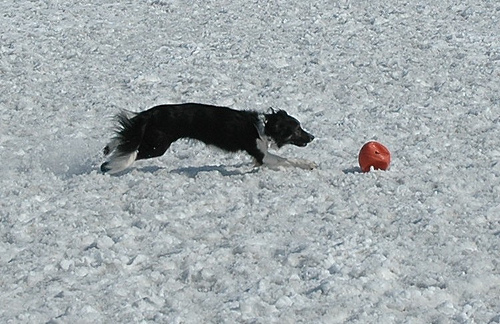}
		\end{minipage}  & \begin{tabular}[c]{@{}l@{}}ein schwarz-weißer hund läuft zu einem\\ kaputten ball im schnee.\\ a black-and-white dog goes for a flattened \\ball on the snow.\end{tabular} & \begin{tabular}[c]{@{}l@{}}\textbf{ours}: a black and white dog runs to a catch \\a ball in the snow.\\ \textbf{supervised:} a black and white dog runs towards \\a broken ball in the snow.\end{tabular} \\ \midrule
		
		\begin{minipage}{.12\textwidth}
			\includegraphics[width=\linewidth]{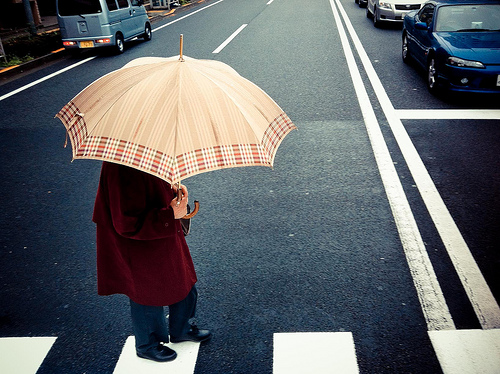}
		\end{minipage} &  \begin{tabular}[c]{@{}l@{}}eine ältere person überquert die straße mit\\ einem regenschirm in der hand.\\an elderly person is crossing a street with an\\ umbrella in their hands. \end{tabular}& \begin{tabular}[c]{@{}l@{}}\textbf{ours}: a elderly man crossing the street with a \\umbrella in the hand.\\ \textbf{supervised:} an elderly person is walking across\\ the street with an umbrella.\end{tabular} \\ \bottomrule
	\end{tabular}
	\caption{Translation examples from our progressive learning approach and the fully supervised NMT model. Images are only for visualization.}
	\label{tab:translation_examples}
\end{table*}

Table~\ref{tab:sota_comparison} presents the performance of different methods for zero-resource machine translation with image pivots.
Since our approach utilizes word translation learned in the first progressive step to benefit sentence translation from two losses, we compare each of the loss and their combination with previous methods.
The $L_{pivot}$ achieves superior performance to previous image-pivoted methods on both datasets and translation directions.
It demonstrates the effectiveness of the re-weighting approach to suppress noises in image-pivoted sentence pairs, while such noises are ignored in previous methods.
The auto-encoding loss $L_{ae}$ brings more significant performance gains, which shows that re-using word embedding matrices from the image-pivoted word translation is effective for the NMT model to encode language-agnostic sentence representation.
The two losses are also complementary with each other and the combination of them achieves the best translation performance on the two datasets.

To be noted, it might be unfair to directly compare our results with the best model in the recent mono-lingual based method  \cite{lample2017unsupervised}.
The best model in \cite{lample2017unsupervised} utilizes word vectors pretrained on a large-scale mono-lingual corpus and multiple iterations of back-translation to achieve good performance on Multi30k dataset (De-En 26.26 and En-De 22.74).
Without pretraining and back-translation, it only achieves translation performance of 7.52 for De-En and 6.24 for En-De on Multi30k dataset.
However, our approach which only employs very limited mono-lingual image caption data and single round training without back-translation can still achieve comparable performance of the best model in \cite{lample2017unsupervised}.
It suggests that image-pivoted approaches could be more effective to translate visually relevant sentences. What is more, our approach is orthogonal to mono-lingual based methods.

\subsection{Ablation Study}

\begin{table}
	\begin{tabular}{l|l} \toprule
		$\beta_i$ & Image-pivoted Pseudo En-De Sentence Pairs \\  \midrule
		0.97 & \begin{tabular}[c]{@{}l@{}}a man in a blue shirt and blue shorts playing tennis .\\ \textcolor{blue}{ein mann in einem blauen oberteil und blauen shorts} \\ \textcolor{blue}{spielt tennis .}\end{tabular} \\ \midrule
		0.70 & \begin{tabular}[c]{@{}l@{}}a brown dog jumps over a fence .\\ \textcolor{blue}{ein} weißer \textcolor{blue}{hund springt über eine} hürde \textcolor{blue}{.}\end{tabular} \\ \midrule
		0.35 & \begin{tabular}[c]{@{}l@{}}a man carving a pumpkin in his boxers .\\ \textcolor{blue}{ein mann in} einem blauen hemd hält einen hammer\textcolor{blue}{.}\end{tabular} \\ \midrule
		0.13 & \begin{tabular}[c]{@{}l@{}}a woman in a white shirt is preparing a meal .\\ peperoni \textcolor{blue}{kochen} im winter !\end{tabular} \\ \bottomrule
	\end{tabular}
	\caption{Image-pivoted pseudo sentence pairs ranked by sentence-level weights. The tokens in the German sentence with high token-level weights are colored in blue. Best viewed in color.}
	\label{tab:weight_examples}
\end{table}

In Table~\ref{tab:ablation_performance}, we evaluate improvements from progressive learning for the two proposed losses.
The loss $L_{pivot}$ without progressive learning denotes utilizing all image-pivoted pseudo pairs for training without re-weighting by the learned word translation.
We can see that it suffers from the noisy pairs and is inferior to our re-weighting model on both datasets.
To gain an intuition on the effects of re-weighting strategy, we illustrate some image-pivoted pseudo sentence pairs in Table~\ref{tab:weight_examples} as well as their sentence and token weights.
The clean pseudo sentence pairs are ranked higher than the noisy ones, and the token-level re-weighting provides a more fine-grained supervision from noisy pairs.
For the loss $L_{ae}$, the lack of progressive learning means we do not employ the learned multi-lingual word representation in the first word translation step.
As shown in Table~\ref{tab:ablation_performance}, the image-pivoted word embedding plays an important role for the effectiveness of auto-encoding loss.
It demonstrates that performance boost from $L_{ae}$ mainly contributes to the progressive learning strategy.

Since the proposed translation learning relies on results from image caption models, we investigate the relation between image caption performance and image-pivoted zero-resource translation performance in Table~\ref{tab:caption_translation_relation}.
Firstly, the translation performance is much higher than the image captioning performance. 
This is mainly because the translation output is more constrained by the input while the caption output is more diverse.
So the image caption model directly used in previous image-pivot translation methods might be a main bottleneck.
Secondly, the translation performance is proportional to the image captioning performance, which indicates that better image captioning can further improve our zero-resource machine translation.


\begin{table}
	\begin{tabular}{c|cc|cc} \toprule
		& \multicolumn{2}{c|}{IAPR-TC12} & \multicolumn{2}{c}{Multi30k} \\
		& I/De-En & I/En-De & I/De-En & I/En-De \\ \midrule
		Caption & 23.0 & 18.9 & 6.6 & 5.6 \\
		Translation & 61.3 & 47.1 & 23.0 & 18.3 \\ \bottomrule
	\end{tabular}
	\caption{The BLEU4 scores of image caption model and image-pivoted translation model on IAPR-TC12 and Multi30k datasets.}
	\label{tab:caption_translation_relation}
\end{table}

\begin{figure} \centering
	\subfigure[IAPR-TC12 dataset.] { \centering
		\label{fig:oracle_cmp_iaprtc12}
		\includegraphics[width=0.47\linewidth]{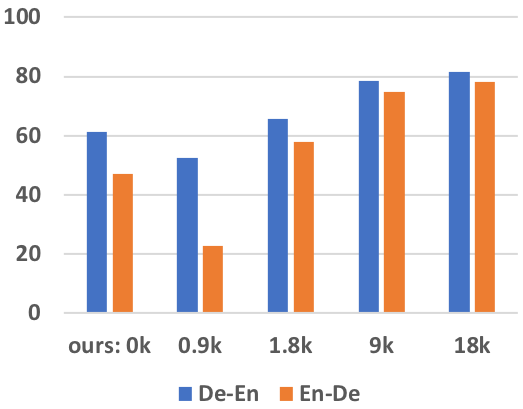}
	} 
	\subfigure[Multi30k dataset.] { \centering
		\label{fig:oracle_cmp_multi30k}
		\includegraphics[width=0.47\linewidth]{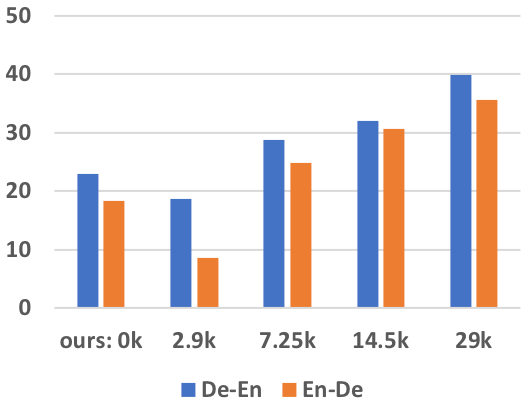}
	}
	\caption{Comparison with the fully supervised NMT model trained on variable amount of parallel texts. The x-axis denotes number of training pairs and y-axis denotes the BLEU4 score.}
	\label{fig:oracle_cmp}
\end{figure}

Finally, we compare our progressively learned zero-resource NMT model with fully supervised NMT model with respect to variable amount of training data in Figure~\ref{fig:oracle_cmp}.
Our proposed model with zero-resource reaches 75\% of the best performance from the supervised NMT model with 18k training pairs for De-to-En, 60\% for En-to-De on IAPR-TC12 dataset. Similarly, it reaches 58\% of the best performance from the supervised NMT model with 29k training pairs for De-to-En and 51\% for En-to-De on Multi30k dataset.
Table~\ref{tab:translation_examples} presents some randomly selected examples.
The proposed approach can generate promising translation results simply based on image pivots without parallel texts.

\section{Conclusion}
In this paper, we address the zero-resource machine translation problem by exploiting visual images as pivots.
Due to the nature that a picture tells a thousand words, description sentences for an image may be semantically nonequivalent, which leads to noisy supervisions to train the NMT model in previous works. 
In this work, we propose a progressive learning approach which consists of progressive easy-to-advanced steps towards learning effective NMT models under zero resource settings.
The learning starts with word-level translation with image pivots and then progresses to sentence-level translation assisted by the word translation and image pivots. 
Experiments on IAPR-TC12 and Multi30k datasets prove the effectiveness of the proposed approach, which significantly outperforms previous image-pivot methods. 

\section*{Acknowledgments}
This work was supported by National Natural Science Foundation of China (No. 61772535), Beijing Natural Science Foundation (No. 4192028), National Key Research and Development Plan (No. 2016YFB1001202) and Research Foundation of Beijing Municipal Science \& Technology Commission (No. Z181100008918002).

\bibliographystyle{named}
\bibliography{ijcai19}

\end{document}